\documentclass[11pt]{article}
\usepackage[final]{acl}
\usepackage{times}
\usepackage{latexsym}
\usepackage[most]{tcolorbox}
\usepackage[T1]{fontenc}
\usepackage[utf8]{inputenc}
\usepackage{microtype}
\usepackage{inconsolata}
\usepackage{multirow} 
\usepackage{graphicx}
\usepackage{booktabs}

\title{Clustered Self-Assessment: A Simple yet Effective Method for \\Uncertainty Quantification in Large Language Models}

\author{
Qi Cao, Takeshi Kojima, Andrew Gambardella,\\
\bf Helinyi Peng, \bf Yutaka Matsuo, \bf Yusuke Iwasawa \\
         The University of Tokyo, Japan\\
         \texttt{\url{qi.cao@weblab.t.u-tokyo.ac.jp}}}

\begin{document}
\maketitle
\begin{abstract}
Large language models (LLMs) demonstrate remarkable performance across diverse tasks, but they often generate responses that appear plausible while being factually incorrect. This problem is compounded by the lack of explicit uncertainty estimates, which makes it difficult for users to judge the reliability of model outputs. Existing uncertainty quantification methods typically rely on indirect signals, such as entropy across sampled generations. These signals can be difficult to interpret and do not fully leverage the model's ability to assess its own uncertainty.
We propose a simple yet effective self-assessment method for uncertainty quantification in LLMs. Our approach groups sampled generations into semantically distinct clusters, converts them into answer options in a structured multiple-choice question, and uses the probability assigned by the LLM to each option as a confidence estimate. Experiments across multiple models and datasets show that our method consistently outperforms baseline approaches. Notably, it achieves competitive performance with as few as two additional samples, demonstrating both its effectiveness and efficiency.\footnote{Code will be available at \url{https://github.com/ccqq77/clustered_self_assessment}.}
\end{abstract}

\section{Introduction}
Large language models (LLMs) have achieved remarkable success across diverse applications. However, ensuring the reliability and factual accuracy of their outputs remains challenging, especially in high-stakes domains such as healthcare, law, and scientific research, where errors can have serious consequences. This issue is compounded by their ability to produce fluent, persuasive text that may obscure inaccuracies and mislead users~\cite{10.1145/3571730,10.1145/3703155}.

To mitigate these risks, it is crucial to develop effective methods for quantifying and communicating uncertainty in LLM-generated outputs. A natural approach is to let LLMs express uncertainty in natural language; however, prior work shows that they often exhibit overconfidence, giving overly confident assessments even when incorrect~\cite{xiong2024can,kadavath2022language}. Researchers have therefore explored alternatives, such as estimating uncertainty from semantic divergence across multiple generations~\cite{kuhn2023semantic,lin2024generating}. While partially effective, these methods primarily measure output discrepancies without leveraging the LLMs' internal understanding of their knowledge~\cite{kadavath2022language}, leading to suboptimal performance. Moreover, many methods rely on indirect uncertainty metrics, such as entropy-based scores~\cite{kuhn2023semantic}, which are difficult for users to interpret and thus limit practical utility~\cite{devic2025calibration}.

Therefore, there is a pressing need for uncertainty quantification methods that leverage the internal capabilities of LLMs while clearly communicating model confidence. To this end, we propose a simple yet effective method that fulfills these requirements. Our main contributions are as follows:

\noindent\textbf{Eliciting self-assessment.} Our approach leverages the internal uncertainty representations of LLMs by using sampled generations to construct structured multiple-choice questions, thereby prompting the model to evaluate its own answers.

\noindent\textbf{Superior performance.} Our method consistently outperforms existing baselines across diverse datasets and models. Notably, it achieves competitive performance with only two additional samples, whereas some baselines require up to sixteen, highlighting its effectiveness.

\noindent\textbf{Human-interpretability.} We use the token probability of each choice as a confidence score, enabling users to intuitively assess the reliability of generated answers. Experimental results show that these confidence scores align well with actual correctness, demonstrating the practicality of our method.

\section{Related Work}

The most relevant studies are sampling-based uncertainty quantification methods, such as Semantic Entropy~\cite{kuhn2023semantic}, EigV, Deg, and Ecc~\cite{lin2024generating}, EigenScore~\cite{chen2024inside}, and SAR~\cite{duan-etal-2024-shifting}. While these methods differ in implementation, they all rely on sampling multiple generations and focus on measuring the discrepancies among sampled answers, without sufficiently leveraging the inherent capabilities of LLMs. This often leads to suboptimal performance. Additionally, the resulting indirect scores are difficult for humans to interpret, which limits their practical utility~\cite{devic2025calibration}.

Another line of work explores prompting LLMs to verbalize their uncertainty~\cite{xiong2024can}. However, due to the well-documented overconfidence issue, such verbalized uncertainty estimates often perform poorly. An alternative approach, P(True)~\cite{kadavath2022language}, queries the LLM about the truthfulness of its own answer and uses the probability of ``True'' as a confidence score, but its performance remains unsatisfactory.

Our method integrates the strengths of sampling-based and self-assessment-based approaches by constructing a structured multiple-choice question for self-assessment, where the options are derived from clustered sampled responses. This design better promotes the model's self-assessment ability and produces calibrated, human-interpretable confidence scores. The relevant methods includes $T^3$~\cite{li-etal-2024-think} and  BSD~\cite{chen-mueller-2024-quantifying}. Compared with $T^3$, our method constructs semantically distinct options by clustering sampled generations, rather than relying on predefined labels. In contrast to BSD, our method derives the confidence score directly from the model's self-assessment, instead of using a weighted aggregation of separate scores.

\section{Methodology}
\label{sec:method}
\begin{figure}[t]
  \centering
    \includegraphics[width=0.5\textwidth]{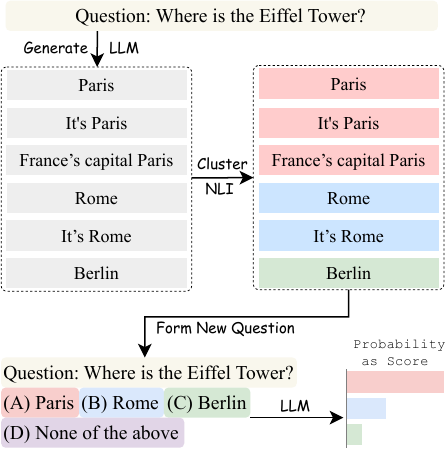}
  \caption{Demonstration of our method using an example from~\citet{farquhar2024detecting}. Sampled generations are clustered based on their semantic compatibility and then used to construct a structured multiple-choice question (MCQ), which is presented as input to the LLM. The probability assigned to each choice serves as its confidence score.}
  \label{fig:framework}
\end{figure}

\begin{figure*}[t]
  \centering
    \includegraphics[width=1.0\textwidth]{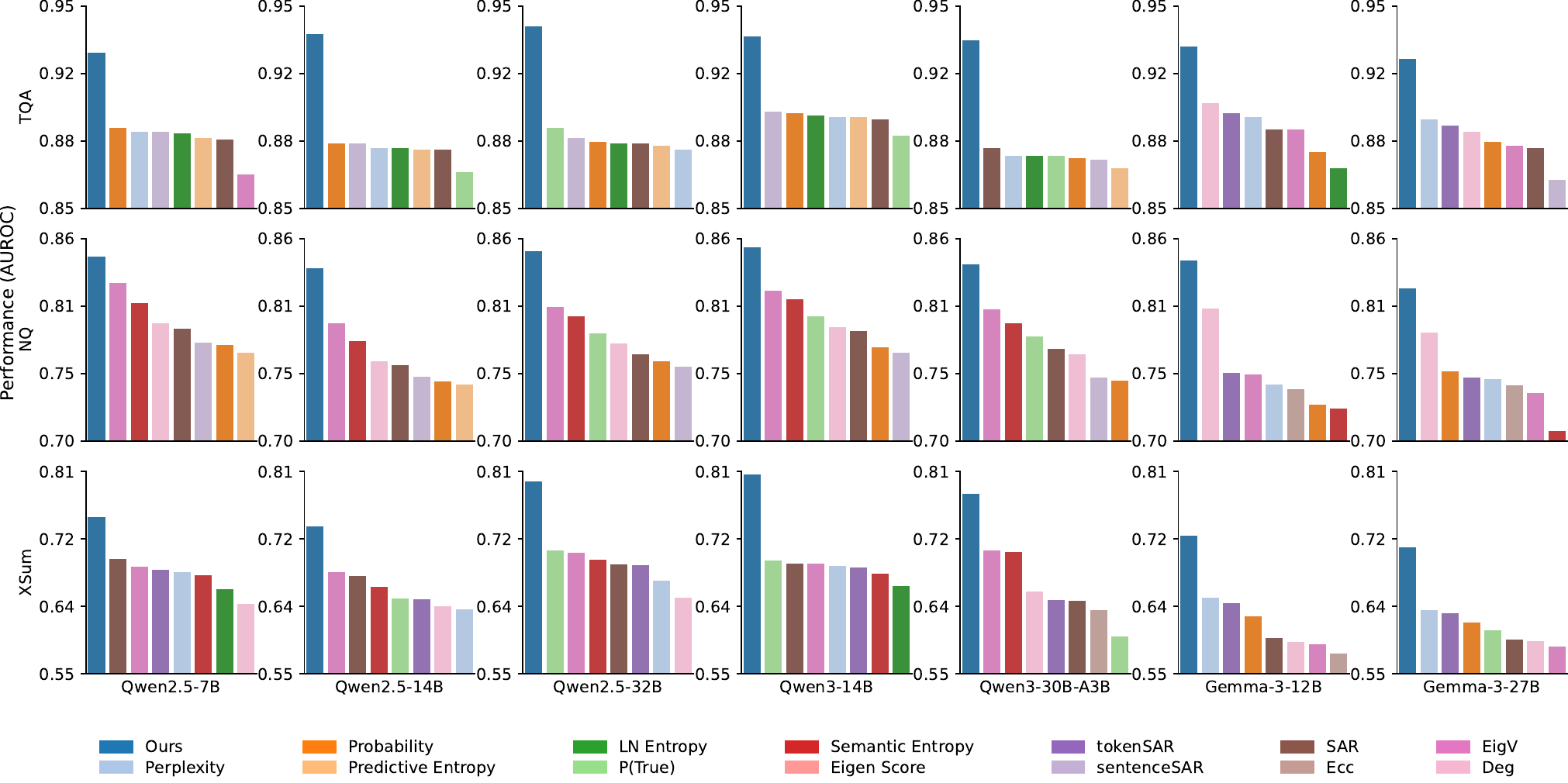}
  \caption{Performance comparison of our method and baseline approaches across different models and datasets. Sampling-based methods use eight additional samples. For clarity, each subfigure shows only the top eight methods ranked by performance, ordered in descending order. Metric: AUROC.}
  \label{fig:performance}
\end{figure*}
As illustrated in Figure~\ref{fig:framework}, our method follows a two-stage process to quantify uncertainty. It first reformulates the question based on the sampled answers and then utilizes the model's own probability estimates as a confidence measure:

\subsection{Answer Clustering} 
For each question, we sample multiple answers from the LLM. Following the answer-clustering procedure introduced in Semantic Entropy~\cite{kuhn2023semantic}, we use an NLI model to estimate pairwise semantic relationships among the sampled answers. Specifically, for each ordered pair $(a_i, a_j)$, the NLI model predicts a label $\ell_{i\to j}\in\{\emph{entailment}, \emph{neutral}, \emph{contradiction}\}$. 
Using the NLI model, we obtain bidirectional pairwise labels for each unordered pair $\{i,j\}$ as the tuple $(\ell_{i\to j}, \ell_{j\to i})$. We then apply a rule-based filter to these bidirectional relationships to determine whether two answers should be grouped together.

In more detail, during clustering, we maintain a set of clusters $\mathcal{G}=\{G_1,\ldots,G_K\}$, each with a representative $r_k \in G_k$ chosen as the first answer assigned to that cluster. We process the answers in a fixed order, beginning with the answer to be assessed, which in our setting is the greedy-decoded answer. For each answer $a_i$, we compare it with the current representatives $\{r_k\}_{k=1}^K$ and assign it to the first cluster for which the representative and $a_i$ satisfy the predefined bidirectional NLI-based grouping criterion. Specifically, two answers are grouped if the bidirectional NLI results contain no \emph{contradiction} label or include at least one \emph{entailment} label in both directions, ensuring that the resulting clusters correspond to semantically distinct answer choices. If no existing cluster satisfies this criterion, we create a new cluster with $a_i$ as its representative.

Finally, we group the answers into semantically compatible clusters, with each cluster corresponding to a distinct candidate answer among the model's sampled generations. This clustering process preserves the diversity of generated answers while reducing semantic redundancy in the sampled outputs, thereby providing a reliable basis for subsequent self-assessment.

\subsection{Self-Assessment} 
We use the clustered answers to construct a structured multiple-choice question (MCQ) with semantically distinct choices, where each choice corresponds to one answer cluster. In addition to the choices derived from the answer clusters, we include an additional option, ``None of the above'', to account for cases where all choices may be incorrect or when the LLM exhibits high uncertainty regarding the correctness of any provided choices. The constructed MCQ is then presented to the original LLM, and the token probability assigned to a specific choice serves as a human-interpretable confidence score.

Formally, let $\mathbf{z} = [z_1, z_2, \ldots, z_V]$ denote the output logits of the LLM over its vocabulary of size $V$, where each logit $z_v$ is the unnormalized score assigned to the $v$-th token. For a specific choice label token $c_i$, the probability assigned by the model is

\begin{equation}
P(c_i) = \frac{\exp(z_{c_i})}{\sum_{v=1}^{V} \exp(z_v)}.
\end{equation}

The confidence score $S$ for the answer to be assessed, associated with label $c_{i^\ast}$ (e.g., corresponding to ``A''), is then explicitly given by
\begin{equation}
S = P(c_{i^\ast}).
\end{equation}

The answer-clustering stage is critical for constructing reliable MCQs for self-assessment. By merging semantically compatible generations, it helps ensure that the resulting choices correspond to distinct candidate answers, preventing compatible alternatives from splitting the model's probability mass and degrading the reliability of uncertainty quantification.

\section{Experiments}

\begin{figure}[b!]
  \centering
    \includegraphics[width=0.5\textwidth]{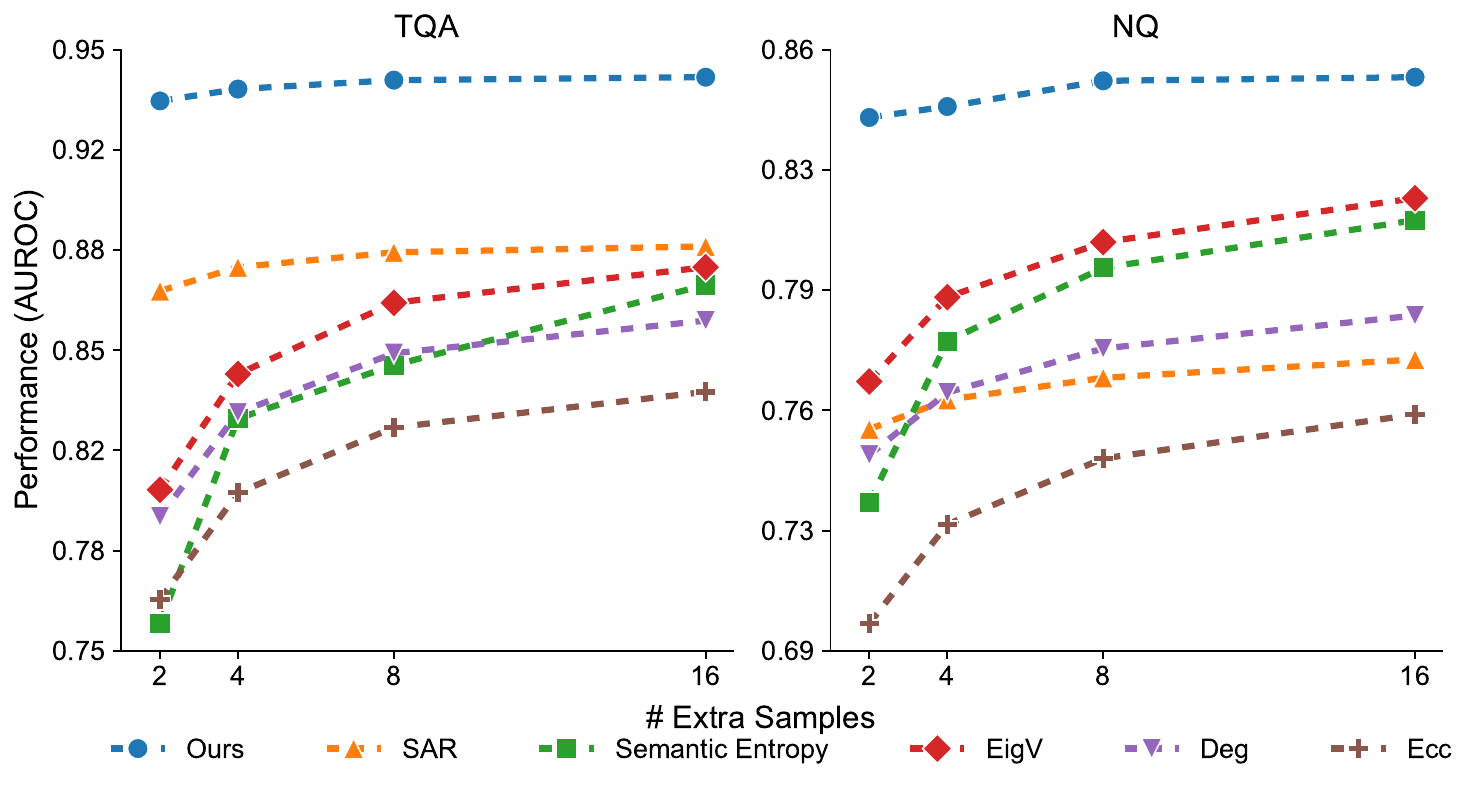}

  \caption{Performance comparison of our method and representative sampling-based baselines using Qwen2.5-32B on TQA and NQ, with varying numbers of additional sampled generations (2–16). Metric: AUROC.}
  \label{fig:samples}
\end{figure}

\begin{table*}[ht]
\centering
\caption{Ablation study evaluating the impact of removing key components, including answer clustering and answer sampling. Metric: AUROC.}
\resizebox{1.0\textwidth}{!}{
\begin{tabular}{llccccccc}
\toprule
Dataset & Method & Qwen2.5-7B & Qwen2.5-14B & Qwen2.5-32B & Qwen3-14B & Qwen3-30B-A3B & Gemma-3-12B & Gemma-3-27B \\
\midrule
\multirow{3}{*}{TQA} & Ours & \textbf{0.927} & \textbf{0.936} & \textbf{0.940} & \textbf{0.935} & \textbf{0.933} & \textbf{0.930} & \textbf{0.924} \\
 & w/o clustering & 0.903 & 0.889 & 0.874 & 0.901 & 0.890 & 0.914 & 0.895 \\
 & w/o sampling & 0.832 & 0.868 & 0.890 & 0.886 & 0.876 & 0.737 & 0.789 \\
\midrule
\multirow{3}{*}{NQ} & Ours & \textbf{0.846} & \textbf{0.837} & \textbf{0.850} & \textbf{0.853} & \textbf{0.840} & \textbf{0.843} & \textbf{0.821} \\
 & w/o clustering & 0.778 & 0.748 & 0.741 & 0.742 & 0.765 & 0.787 & 0.766 \\
 & w/o sampling & 0.729 & 0.731 & 0.785 & 0.799 & 0.783 & 0.609 & 0.659 \\
\bottomrule
\end{tabular}
}
\label{table:ablation}
\end{table*}

\subsection{Setup}
\paragraph{Datasets.} We use two commonly adopted datasets for uncertainty quantification in our main experiments and further analysis: TriviaQA (TQA;~\citealp{joshi-etal-2017-triviaqa}) and Natural Questions (NQ;~\citealp{kwiatkowski-etal-2019-natural}). For preprocessing, we follow the approach described in~\citet{lin2024generating}, resulting in 9,960 samples for TQA and 3,610 samples for NQ. In addition to these two standard question answering (QA) datasets, we include Extreme Summarization (XSum;~\citealp{narayan-etal-2018-dont}) as a supplementary dataset for summarization, allowing us to evaluate performance on longer responses. Following the LM-Polygraph Benchmark~\cite{vashurin-etal-2025-benchmarking}, we use the 11,334 samples from the XSum test set.

\paragraph{Models.} We use seven state-of-the-art open-source models from different series and with varying sizes: the 7B, 14B, and 32B models from the Qwen-2.5 series~\cite{qwen2025qwen25technicalreport}; the 14B and 30B-A3B models from the Qwen-3 series~\cite{yang2025qwen3technicalreport}; and the 12B and 27B models from the Gemma-3 series~\cite{gemmateam2025gemma3technicalreport}. To ensure comparability, we use the base version for the Qwen-3 series and the pt version for the Gemma-3 series, and omit the suffixes for brevity.

\paragraph{Baselines.} We compare our method against a range of existing uncertainty quantification approaches, including Perplexity~\cite{fomicheva-etal-2020-unsupervised}, Probability (the probability of the generated sequence), P(True)~\citep{kadavath2022language}, Predictive Entropy~\citep{lindley1956measure}, LN-Entropy~\cite{malinin2021uncertainty}, Semantic Entropy~\cite{kuhn2023semantic}, as well as Ecc, EigV, and Deg~\cite{lin2024generating}, and tokenSAR, sentenceSAR, and SAR~\cite{duan-etal-2024-shifting}.

The details about implementation can be found in Appendix~\ref{sec:implementations}.

\subsection{Performance}
As shown in Figure~\ref{fig:performance}, we compare our uncertainty quantification method with various baseline approaches. For the sampling-based approaches, including our method and the corresponding baselines, we use eight additional samples. We evaluate performance on two standard QA datasets, TQA and NQ, as the main datasets, and additionally on the XSum dataset for the text summarization task, which features longer responses. Across all settings, our method consistently and substantially outperforms the baselines, demonstrating its effectiveness across both QA and summarization tasks.

\subsection{Sample Efficiency}
Figure~\ref{fig:samples} compares our method with representative sampling-based approaches under varying numbers of additional samples ($ n \in \{2, 4, 8, 16\}$) using Qwen2.5-32B on TQA and NQ datasets. Detailed results are provided in Appendix~\ref{sec:detailed_results}. While most methods show improved performance as the number of samples increases, our approach consistently outperforms the baselines across all settings and achieves strong results even with a small number of samples. Notably, our approach achieves competitive performance with only two additional samples, whereas some baselines require up to sixteen. Since most of the computational overhead arises from generating additional samples, whereas the NLI model is lightweight and obtaining logits for the choices requires only a single token generation, our method achieves competitive performance with substantially lower computational overhead.

\begin{table*}[ht]
\centering
\caption{Calibration comparison of our method and two probability-based baselines. Metric: Brier score.}
\resizebox{1.0\textwidth}{!}{
\begin{tabular}{llccccccc}
\toprule
Dataset & Method & Qwen2.5-7B & Qwen2.5-14B & Qwen2.5-32B & Qwen3-14B & Qwen3-30B-A3B & Gemma-3-12B & Gemma-3-27B \\
\midrule
\multirow{4}{*}{TQA} & Ours & \textbf{0.1183} & \textbf{0.0917} & \textbf{0.0843} & \textbf{0.0921} & \textbf{0.0894} & \textbf{0.0943} & \textbf{0.0721} \\
 & P(True) & 0.1708 & 0.1400 & 0.1172 & 0.1315 & 0.1398 & 0.2147 & 0.1758 \\
 & Probability & 0.1674 & 0.2001 & 0.2267 & 0.1807 & 0.1917 & 0.2402 & 0.1975 \\
 & NSE & 0.1580 & 0.1342 & 0.1200 & 0.1283 & 0.1289 & 0.1421 & 0.0937 \\
\midrule
\multirow{4}{*}{NQ} & Ours & \textbf{0.1706} & \textbf{0.1729} & \textbf{0.1597} & \textbf{0.1611} & \textbf{0.1656} & \textbf{0.1809} & \textbf{0.1736} \\
 & P(True) & 0.1993 & 0.2128 & 0.1918 & 0.2015 & 0.2191 & 0.2462 & 0.2354 \\
 & Probability & 0.2471 & 0.3409 & 0.3155 & 0.3149 & 0.3208 & 0.3207 & 0.3503 \\
 & NSE & 0.2007 & 0.2205 & 0.1993 & 0.1930 & 0.2011 & 0.2581 & 0.2462 \\
\bottomrule
\end{tabular}
}
\label{table:calibration}
\end{table*}

\subsection{Ablation Study}

In this ablation study, we examine the impact of removing two key components, answer clustering and answer sampling, to evaluate their individual contributions to overall performance. When clustering is removed, the raw sampled generations are directly used as separate choices. When sampling is removed, the method degenerates to P(True), as only a single choice remains available. Table~\ref{table:ablation} presents the results of this analysis. Removing either component leads to a substantial decline in performance, underscoring the importance of both answer clustering and answer sampling in our approach. Specifically, answer sampling enables the model to consider a diverse set of candidate answers for comparison, while answer clustering mitigates confusion caused by semantically compatible choices.

\begin{figure}[!b]
  \centering
    \includegraphics[width=0.5\textwidth]{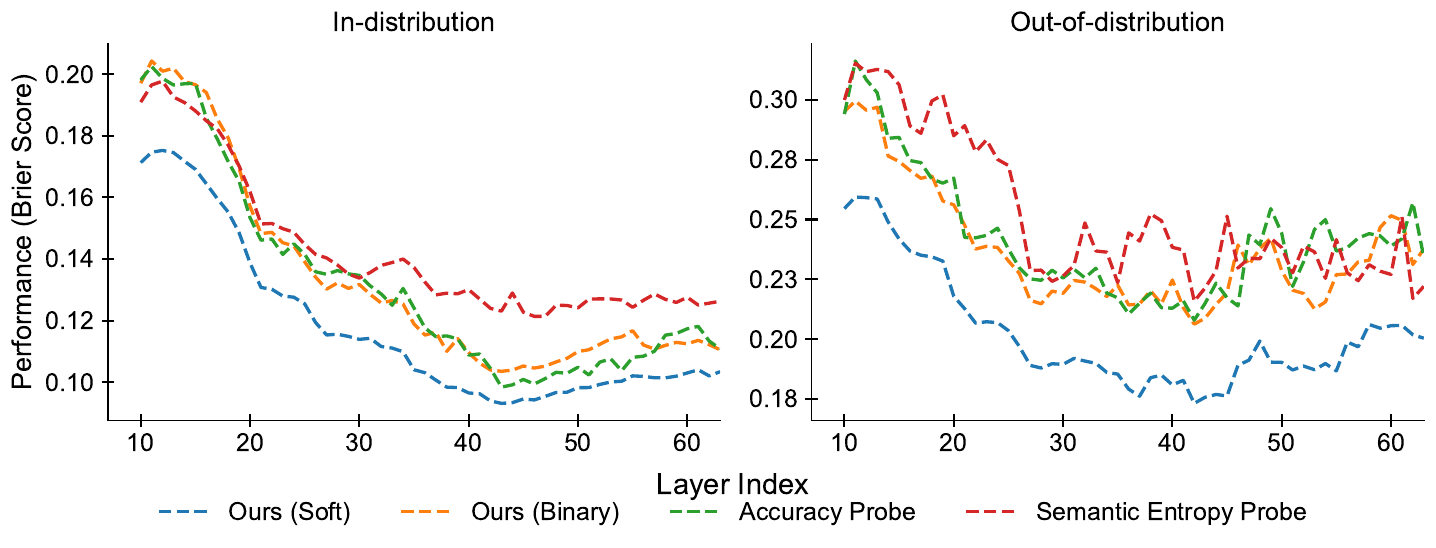}

  \caption{Performance comparison of our probes and baselines using Qwen2.5-32B under in-distribution and out-of-distribution settings. Metric: Brier score.}
  \label{fig:probe_results}
\end{figure}

\subsection{Calibration}

We further examine calibration to determine whether the confidence scores produced by our method are aligned with empirical correctness, which is essential for reliable and actionable uncertainty estimation. We compare our approach with two probability-based baselines: P(True) and Probability. In addition to these probability-based baselines, we include a normalized semantic entropy (NSE) baseline, where the semantic entropy is scaled to the range [0, 1] based on the number of semantic clusters in the sampled responses. As shown in Table~\ref{table:calibration}, we report results using the Brier score to measure the calibration of uncertainty scores against correctness. Across all settings, our method consistently outperforms the baselines, demonstrating superior calibration and greater practical utility.

\subsection{Probe}

In prior work, Semantic Entropy Probe (SEP)~\cite{kossen2024semanticentropyprobesrobust} trains a probe to predict uncertainty from LLM hidden states using semantic entropy scores, thereby eliminating the need for multiple samples and improving efficiency. Similarly, we use our score as the training signal for the probe. We consider two settings: (1) binarizing the score with a threshold of 0.5, and (2) using it directly as a soft label. Figure~\ref{fig:probe_results} compares our probes against SEP and an accuracy probe trained on correctness labels, using hidden states from intermediate layers in both in-distribution and out-of-distribution settings (implementation details are provided in Appendix~\ref{sec:probe_train}, and full results are shown in Appendix~\ref{sec:detailed_results}). Our soft-label probe achieves performance comparable to that of the accuracy probe in the in-distribution setting and substantially outperforms the baselines in the out-of-distribution setting, without requiring correctness labels. These results suggest that our probe is both practical and reliable, and that our score more closely aligns with the internal uncertainty representations of LLMs.

\section{Conclusion}

We present \textit{Clustered Self-Assessment}, a simple yet effective method for uncertainty quantification that enhances LLM self-assessment by clustering sampled answers to construct MCQs with semantically distinct choices. Experiments across diverse models and datasets show that our approach consistently outperforms existing baselines. Further analysis demonstrates that the resulting confidence scores are well calibrated and can be used to train a probe that predicts uncertainty from LLM hidden states, highlighting the effectiveness and practicality of our method.

\section{Limitations}

Our method has several limitations that suggest directions for future improvement. 

First, it requires access to output logits to compute the uncertainty score. Such access is not always available, particularly for closed-source models, which limits the applicability of our approach in certain deployment scenarios.

Second, our method depends on an external NLI model to cluster generated answers. This reliance introduces additional computational overhead and reduces self-containment. Moreover, the external NLI model may struggle to capture nuanced semantics or handle out-of-distribution cases effectively. A promising direction for future work is to leverage the internal representations of the LLM itself for answer clustering, as LLMs typically encode richer semantic information than smaller NLI models. Integrating this capability could improve both computational efficiency and the robustness of uncertainty quantification.

Finally, our method directly uses the LLM's softmax probabilities as confidence scores without applying post-hoc calibration. While this design already achieves strong performance, incorporating a calibration step could further enhance the reliability and interpretability of the confidence estimates, leading to more consistent and trustworthy results.
\bibliography{custom,anthology-1,anthology-2}

\appendix

\section{Implementation Details}

\subsection{Main Experiments}
\label{sec:implementations}
We benchmark our uncertainty quantification method in a deterministic setting, where LLMs generate responses using greedy decoding. This setup reflects practical applications, such as factual question answering, where producing reliable outputs is essential. Specifically, our objective is to quantify the uncertainty associated with generations produced by greedy decoding. For sampling-based methods, we estimate uncertainty by generating additional responses with temperature sampling ($\tau=0.5$, $\mathrm{top}\text{-}k=32$, and $\mathrm{top}\text{-}p=0.95$), yielding 2 to 16 supplementary generations per question. This temperature-sampling configuration follows prior work showing that similar settings are effective for uncertainty quantification methods, including EigenScore~\cite{chen2024inside} and Semantic Entropy~\cite{kuhn2023semantic}.

For answer clustering, our method employs \texttt{deberta-large-mnli}~\cite{he2021deberta} as the default external NLI model. The same model is also used by Semantic Entropy~\cite{kuhn2023semantic} and by EigV, Deg, and Ecc~\cite{lin2024generating}. In contrast, SAR~\cite{duan-etal-2024-shifting} follows its default configuration and utilizes \texttt{stsb-roberta-large}~\cite{reimers-gurevych-2019-sentence} to compute sentence similarity.

To establish ground truth for uncertainty evaluation, we assess the correctness of each generated answer. Traditional metrics such as ROUGE-L~\cite{lin-2004-rouge} and sentence-level semantic similarity~\cite{reimers-gurevych-2019-sentence} can be unreliable due to incomplete reference answers, arbitrary thresholds, and spurious interactions between uncertainty scores and evaluation metrics~\cite{santilli2024on}. To address these issues, we employ GPT-4.1 (2025-04-14 version)~\cite{openai2025gpt41} as an automatic judge. GPT-4.1 is prompted to directly evaluate the factual accuracy of each response, producing reliable correctness labels that mitigate the limitations of heuristic-based evaluation. Using these labels, we compute the area under the receiver operating characteristic curve (AUROC) to quantify how effectively each uncertainty estimation method distinguishes correct from incorrect answers.

\subsection{Training Probe Classifier}
\label{sec:probe_train}
Following the practice of SEP~\cite{kossen2024semanticentropyprobesrobust}, we train a probe to predict uncertainty scores from the LLM hidden states. The probe is trained using 70\% of the TQA samples as training data, as TQA provides a larger dataset. For the in-distribution setting, the remaining 30\% of TQA samples are used for testing, while for the out-of-distribution setting, NQ samples serve as the test data. To ensure a fair comparison, we use scores computed with 16 additional samples for both SEP and our probes. A simple logistic regression classifier is employed as the probe, using the hidden states of the second-to-last tokens in the generated answers as input features.

For the accuracy probe, the correctness labels obtained from the main experiments are used as supervision. For SEP, the semantic entropy scores from the main experiments are binarized using the optimal split defined in its implementation. For our method, we evaluate two configurations: one using a binarized label with a threshold of 0.5, and another using the raw score directly as a soft label.

In our experiments, we use a logistic regression classifier implemented in scikit-learn~\cite{scikit-learn} as the probing model. Since hidden states are high-dimensional and many dimensions may be irrelevant to uncertainty representations, we apply $\ell_1$ regularization with $C=0.1$ to reduce redundancy and mitigate spurious correlations.

\section{Additional Analysis}
\subsection{Sensitivity Analysis}
To assess the robustness of our model under different settings, we perform a comprehensive sensitivity analysis along three complementary axes, while keeping all other components of the pipeline fixed to their default configuration.

\begin{itemize}
    
    \item \textbf{Answer order in MCQ.} We analyze the effect of the order in which candidate answers are presented in the MCQ, comparing the default \textit{Original} ordering with two alternatives: \textit{Reverse} and \textit{Random}. The corresponding results are presented in Table~\ref{table:order_auroc}. 

    \item \textbf{NLI model scale.} We examine the impact of model size by evaluating three NLI backbones: \texttt{deberta-large-mnli} (v1-large) as the default configuration, along with the larger models \texttt{deberta-v2-xlarge-mnli} (v2-xlarge) and \texttt{deberta-v2-xxlarge-mnli} (v2-xxlarge). The corresponding results are shown in Table~\ref{table:nli_auroc}.

    \item \textbf{Sampling temperature.} We vary the sampling temperature used to generate additional answers, sweeping over $\tau \in \{0.25, 0.5, 0.75, 1.0\}$. The corresponding results are presented in Table~\ref{table:temp_auroc}. 

\end{itemize}

All other experimental settings follow the evaluation protocol described in Appendix~\ref{sec:implementations}, reporting AUROC on NQ and TQA across different model configurations, with the number of additional samples fixed at eight. Across most settings, our method shows only marginal performance variation, demonstrating its robustness and stability under different configurations.

\begin{table*}[!t]
\centering
\caption{Sensitivity analysis of answer order. Metric: AUROC.}
\label{table:order_auroc}
\resizebox{\textwidth}{!}{
\begin{tabular}{llccccccc}
\toprule
Dataset & Order & Qwen2.5-7B & Qwen2.5-14B & Qwen2.5-32B & Qwen3-14B & Qwen3-30B-A3B & Gemma-3-12B & Gemma-3-27B \\
\midrule
\multirow{3}{*}{TQA} & Original & 0.927 & 0.936 & 0.940 & 0.935 & 0.933 & 0.930 & 0.924 \\
 & Reverse & 0.921 & 0.933 & 0.938 & 0.933 & 0.929 & 0.933 & 0.925 \\
 & Random & 0.924 & 0.935 & 0.939 & 0.934 & 0.930 & 0.931 & 0.923 \\
\midrule
\multirow{3}{*}{NQ} & Original & 0.846 & 0.837 & 0.850 & 0.853 & 0.840 & 0.843 & 0.821 \\
 & Reverse & 0.837 & 0.835 & 0.852 & 0.855 & 0.841 & 0.822 & 0.797 \\
 & Random & 0.843 & 0.837 & 0.848 & 0.857 & 0.840 & 0.828 & 0.809 \\
\bottomrule
\end{tabular}
}
\end{table*}

\begin{table*}[!t]
\centering
\caption{Sensitivity analysis of NLI model. Metric: AUROC.}
\label{table:nli_auroc}
\resizebox{\textwidth}{!}{
\begin{tabular}{llccccccc}
\toprule
Dataset & NLI Model & Qwen2.5-7B & Qwen2.5-14B & Qwen2.5-32B & Qwen3-14B & Qwen3-30B-A3B & Gemma-3-12B & Gemma-3-27B \\
\midrule
\multirow{3}{*}{TQA} & v1-large & 0.927 & 0.936 & 0.940 & 0.935 & 0.933 & 0.930 & 0.924 \\
 & v2-xlarge & 0.928 & 0.936 & 0.940 & 0.935 & 0.934 & 0.929 & 0.925 \\
 & v2-xxlarge & 0.928 & 0.936 & 0.940 & 0.936 & 0.934 & 0.930 & 0.924 \\
\midrule
\multirow{3}{*}{NQ} & v1-large & 0.846 & 0.837 & 0.850 & 0.853 & 0.840 & 0.843 & 0.821 \\
 & v2-xlarge & 0.853 & 0.837 & 0.848 & 0.852 & 0.839 & 0.840 & 0.816 \\
 & v2-xxlarge & 0.851 & 0.838 & 0.849 & 0.854 & 0.841 & 0.843 & 0.822 \\
\bottomrule
\end{tabular}
}
\end{table*}

\begin{table*}[!t]
\centering
\caption{Sensitivity analysis of sampling temperature. Metric: AUROC.}
\label{table:temp_auroc}
\resizebox{\textwidth}{!}{
\begin{tabular}{llccccccc}
\toprule
Dataset & Temperature & Qwen2.5-7B & Qwen2.5-14B & Qwen2.5-32B & Qwen3-14B & Qwen3-30B-A3B & Gemma-3-12B & Gemma-3-27B \\
\midrule
\multirow{4}{*}{TQA} & 0.25 & 0.917 & 0.932 & 0.934 & 0.930 & 0.925 & 0.925 & 0.920 \\
 & 0.5 & 0.927 & 0.936 & 0.940 & 0.935 & 0.933 & 0.930 & 0.924 \\
 & 0.75 & 0.928 & 0.938 & 0.938 & 0.936 & 0.933 & 0.930 & 0.927 \\
 & 1.0 & 0.926 & 0.935 & 0.937 & 0.931 & 0.929 & 0.926 & 0.926 \\
\midrule
\multirow{4}{*}{NQ} & 0.25 & 0.835 & 0.835 & 0.848 & 0.854 & 0.839 & 0.840 & 0.825 \\
 & 0.5 & 0.846 & 0.837 & 0.850 & 0.853 & 0.840 & 0.843 & 0.821 \\
 & 0.75 & 0.841 & 0.835 & 0.841 & 0.841 & 0.832 & 0.835 & 0.818 \\
 & 1.0 & 0.835 & 0.826 & 0.837 & 0.841 & 0.827 & 0.827 & 0.808 \\
\bottomrule
\end{tabular}
}
\end{table*}
\subsection{Embedding-based Alternatives for Answer Clustering}

As an alternative to the NLI-based answer clustering module, we investigate embedding-based approaches by replacing the NLI model with the following embedding types:
\begin{itemize}
    \item \textbf{OpenAI}: Embeddings from OpenAI's \texttt{text-embedding-3-large}\footnote{\url{https://developers.openai.com/api/docs/models/text-embedding-3-large}};
    \item \textbf{MPNet}: Embeddings from the open-source sentence embedding model
    \texttt{paraphrase\allowbreak-multilingual\allowbreak-mpnet\allowbreak-base\allowbreak-v2}~\cite{reimers-gurevych-2019-sentence};
    \item \textbf{Self}: Embeddings from the LLM's hidden states, extracted from the middle layer at the second-to-last token during generation.
\end{itemize}

We evaluate two clustering strategies:
\begin{itemize}
    \item \textbf{Threshold}: Answer embeddings are grouped based on cosine similarity using predefined thresholds of 0.2, 0.4, 0.6, and 0.8.
    \item \textbf{K-means}: Answer embeddings are partitioned into 2, 3, or 4 clusters using the K-means algorithm~\cite{macqueen1967multivariate}.
\end{itemize}

The results are reported in Table~\ref{table:emb_auroc}. Overall, the NLI-based clustering method outperforms the embedding-based alternatives in most configurations, suggesting that explicit entailment modeling is more effective for answer clustering in our proposed framework.

\section{Detailed Experimental Results}
\label{sec:detailed_results}
\begin{figure*}[htbp]
  \centering
  \includegraphics[width=1.0\textwidth]{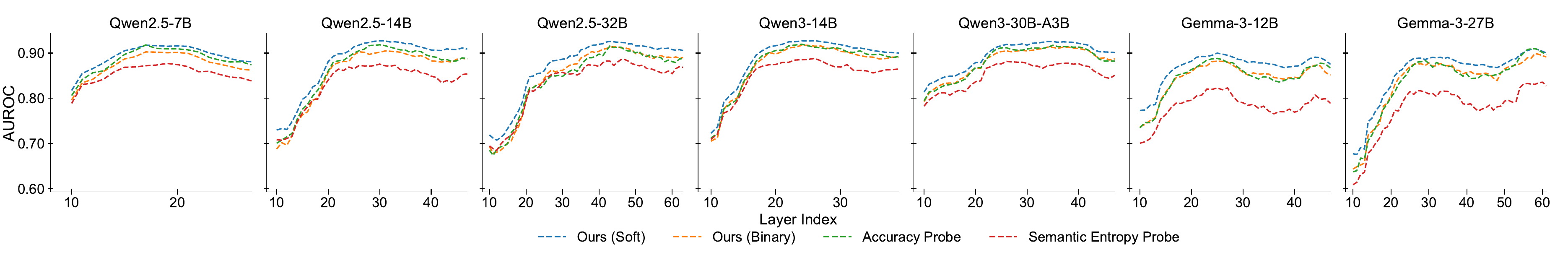}
  \includegraphics[width=1.0\textwidth]{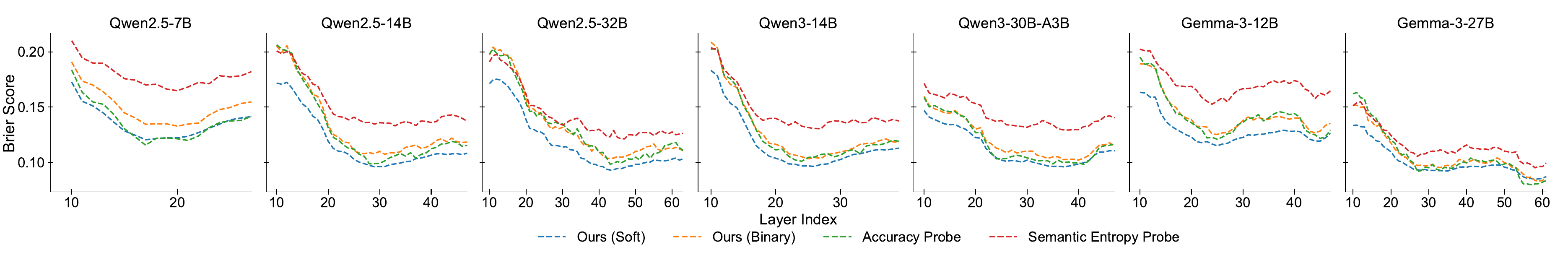}
  
  \includegraphics[width=1.0\textwidth]{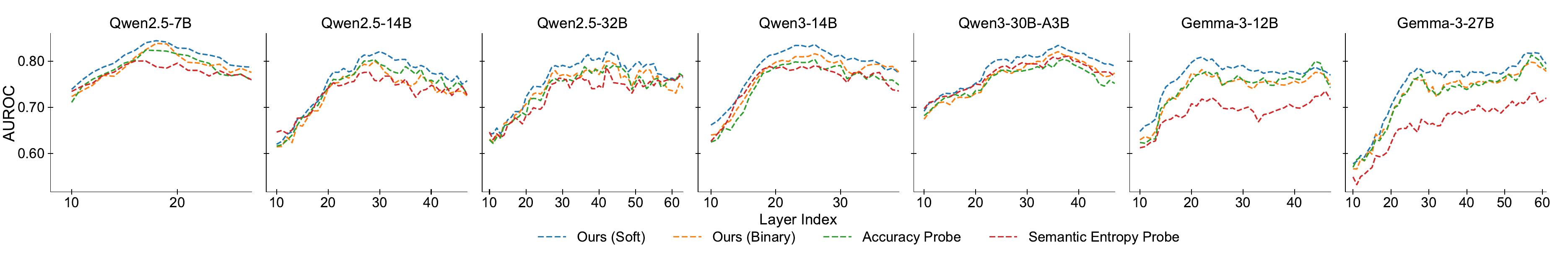}
  \includegraphics[width=1.0\textwidth]{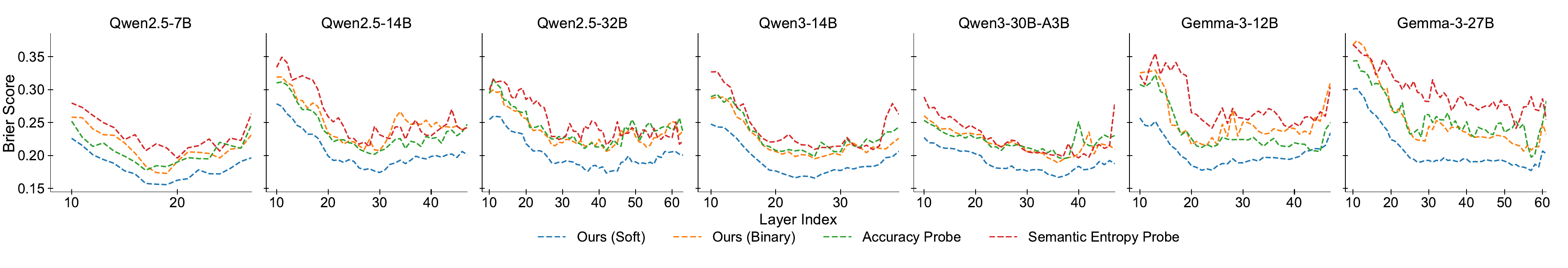}
  
  \caption{Experimental results for probes that predict scores from hidden states in intermediate layers. First row: In-distribution results evaluated using AUROC. Second row: In-distribution results evaluated using the Brier score. Third row: Out-of-distribution results evaluated using AUROC. Fourth row: Out-of-distribution results evaluated using the Brier score.}
  \label{fig:probe_full}
\end{figure*}

The detailed experimental results across different models and methods with varying numbers of additional samples (2, 4, 8, and 16) on the TQA and NQ datasets are presented in Table~\ref{table:performance_plus_2}, Table~\ref{table:performance_plus_4}, Table~\ref{table:performance_plus_8}, and Table~\ref{table:performance_plus_16}, respectively. The experimental results on the XSum dataset with eight additional samples are shown in Table~\ref{table:performance_xsum_8}. The complete probe results are provided in Figure~\ref{fig:probe_full}.

\begin{table*}[htbp]
\centering
\small
\setlength{\tabcolsep}{4pt}
\renewcommand{\arraystretch}{1.08}
\caption{Comparison of embedding-based clustering alternatives with the NLI-based clustering method. Metric: AUROC.}
\label{table:emb_auroc}
\resizebox{\textwidth}{!}{
\begin{tabular}{@{}llllccccccc@{}}
\toprule
Dataset & Embedding & Criterion & Param & Qwen2.5-7B & Qwen2.5-14B & Qwen2.5-32B & Qwen3-14B & Qwen3-30B-A3B & Gemma-3-12B & Gemma-3-27B \\
\midrule
\multirow{22}{*}{TQA} & \multirow{7}{*}{OpenAI} & \multirow{4}{*}{Threshold} & 0.2 & 0.897 & 0.921 & 0.923 & 0.916 & 0.908 & 0.881 & 0.887 \\
 &  &  & 0.4 & 0.915 & 0.928 & 0.931 & 0.925 & 0.921 & 0.904 & 0.903 \\
 &  &  & 0.6 & \underline{0.926} & \underline{0.931} & 0.933 & 0.931 & 0.925 & 0.929 & \textbf{0.924} \\
 &  &  & 0.8 & 0.921 & 0.915 & 0.919 & 0.924 & 0.918 & \underline{0.930} & \underline{0.922} \\
\cmidrule(l){3-11}
 &  & \multirow{3}{*}{K-means} & 2 & 0.878 & 0.877 & 0.865 & 0.887 & 0.881 & 0.893 & 0.876 \\
 &  &  & 3 & 0.895 & 0.882 & 0.865 & 0.893 & 0.885 & 0.908 & 0.888 \\
 &  &  & 4 & 0.900 & 0.885 & 0.868 & 0.896 & 0.887 & 0.913 & 0.893 \\
\cmidrule(l){2-11}
 & \multirow{7}{*}{MPNet} & \multirow{4}{*}{Threshold} & 0.2 & 0.900 & 0.921 & 0.925 & 0.918 & 0.911 & 0.887 & 0.891 \\
 &  &  & 0.4 & 0.916 & 0.930 & 0.933 & 0.929 & 0.925 & 0.911 & 0.910 \\
 &  &  & 0.6 & \underline{0.926} & \textbf{0.936} & \underline{0.936} & \underline{0.933} & \underline{0.928} & 0.928 & \underline{0.922} \\
 &  &  & 0.8 & 0.924 & 0.926 & 0.927 & 0.929 & 0.923 & \textbf{0.932} & \textbf{0.924} \\
\cmidrule(l){3-11}
 &  & \multirow{3}{*}{K-means} & 2 & 0.880 & 0.878 & 0.867 & 0.888 & 0.882 & 0.893 & 0.876 \\
 &  &  & 3 & 0.896 & 0.882 & 0.865 & 0.894 & 0.885 & 0.909 & 0.889 \\
 &  &  & 4 & 0.900 & 0.885 & 0.868 & 0.897 & 0.887 & 0.913 & 0.893 \\
\cmidrule(l){2-11}
 & \multirow{7}{*}{Self} & \multirow{4}{*}{Threshold} & 0.2 & 0.896 & 0.921 & 0.923 & 0.915 & 0.908 & 0.879 & 0.886 \\
 &  &  & 0.4 & 0.896 & 0.921 & 0.922 & 0.915 & 0.908 & 0.879 & 0.886 \\
 &  &  & 0.6 & 0.905 & 0.917 & 0.923 & 0.915 & 0.912 & 0.879 & 0.886 \\
 &  &  & 0.8 & 0.921 & 0.920 & 0.921 & 0.923 & 0.918 & 0.879 & 0.886 \\
\cmidrule(l){3-11}
 &  & \multirow{3}{*}{K-means} & 2 & 0.878 & 0.877 & 0.863 & 0.887 & 0.882 & 0.893 & 0.876 \\
 &  &  & 3 & 0.895 & 0.882 & 0.865 & 0.893 & 0.885 & 0.909 & 0.888 \\
 &  &  & 4 & 0.900 & 0.885 & 0.867 & 0.896 & 0.886 & 0.913 & 0.893 \\
\cmidrule(l){2-11}
 & \multicolumn{3}{l}{NLI} & \textbf{0.927} & \textbf{0.936} & \textbf{0.940} & \textbf{0.935} & \textbf{0.933} & \underline{0.930} & \textbf{0.924} \\
\midrule
\multirow{22}{*}{NQ} & \multirow{7}{*}{OpenAI} & \multirow{4}{*}{Threshold} & 0.2 & 0.804 & 0.820 & 0.827 & 0.840 & 0.823 & 0.807 & 0.791 \\
 &  &  & 0.4 & 0.818 & \underline{0.822} & \underline{0.833} & 0.837 & 0.826 & 0.819 & 0.782 \\
 &  &  & 0.6 & 0.817 & 0.798 & 0.808 & 0.817 & 0.808 & \underline{0.825} & \underline{0.800} \\
 &  &  & 0.8 & 0.800 & 0.776 & 0.777 & 0.785 & 0.787 & 0.812 & 0.795 \\
\cmidrule(l){3-11}
 &  & \multirow{3}{*}{K-means} & 2 & 0.761 & 0.743 & 0.730 & 0.752 & 0.764 & 0.789 & 0.764 \\
 &  &  & 3 & 0.765 & 0.735 & 0.725 & 0.731 & 0.753 & 0.790 & 0.752 \\
 &  &  & 4 & 0.769 & 0.732 & 0.725 & 0.726 & 0.752 & 0.788 & 0.758 \\
\cmidrule(l){2-11}
 & \multirow{7}{*}{MPNet} & \multirow{4}{*}{Threshold} & 0.2 & 0.809 & 0.820 & 0.828 & 0.838 & 0.822 & 0.809 & 0.797 \\
 &  &  & 0.4 & \underline{0.826} & \underline{0.822} & \underline{0.833} & 0.838 & \underline{0.827} & 0.820 & 0.791 \\
 &  &  & 0.6 & 0.821 & 0.811 & 0.823 & 0.824 & 0.820 & 0.823 & 0.793 \\
 &  &  & 0.8 & 0.813 & 0.794 & 0.800 & 0.804 & 0.805 & 0.819 & 0.797 \\
\cmidrule(l){3-11}
 &  & \multirow{3}{*}{K-means} & 2 & 0.759 & 0.744 & 0.734 & 0.754 & 0.763 & 0.791 & 0.765 \\
 &  &  & 3 & 0.765 & 0.733 & 0.724 & 0.736 & 0.753 & 0.790 & 0.753 \\
 &  &  & 4 & 0.770 & 0.734 & 0.723 & 0.728 & 0.751 & 0.789 & 0.759 \\
\cmidrule(l){2-11}
 & \multirow{7}{*}{Self} & \multirow{4}{*}{Threshold} & 0.2 & 0.805 & 0.821 & 0.826 & \underline{0.841} & 0.822 & 0.808 & \underline{0.800} \\
 &  &  & 0.4 & 0.802 & 0.820 & 0.817 & 0.839 & 0.821 & 0.808 & \underline{0.800} \\
 &  &  & 0.6 & 0.781 & 0.779 & 0.782 & 0.803 & 0.798 & 0.808 & \underline{0.800} \\
 &  &  & 0.8 & 0.783 & 0.761 & 0.775 & 0.782 & 0.774 & 0.808 & \underline{0.800} \\
\cmidrule(l){3-11}
 &  & \multirow{3}{*}{K-means} & 2 & 0.757 & 0.738 & 0.732 & 0.750 & 0.754 & 0.792 & 0.763 \\
 &  &  & 3 & 0.766 & 0.736 & 0.720 & 0.731 & 0.751 & 0.788 & 0.753 \\
 &  &  & 4 & 0.771 & 0.734 & 0.721 & 0.726 & 0.750 & 0.787 & 0.758 \\
\cmidrule(l){2-11}
 & \multicolumn{3}{l}{NLI} & \textbf{0.846} & \textbf{0.837} & \textbf{0.850} & \textbf{0.853} & \textbf{0.840} & \textbf{0.843} & \textbf{0.821} \\
\bottomrule
\end{tabular}
}
\end{table*}

\begin{table*}[htbp]
\centering\caption{Experimental results on TQA and NQ with 2 additional samples. Metric: AUROC.}
\resizebox{\textwidth}{!}{
\begin{tabular}{llccccccc}
\toprule
Dataset & Method & Qwen2.5-7B & Qwen2.5-14B & Qwen2.5-32B & Qwen3-14B & Qwen3-30B-A3B & Gemma-3-12B & Gemma-3-27B \\
\midrule
\multirow{14}{*}{TQA} & Perplexity & 0.888 & 0.880 & 0.879 & 0.895 & \underline{0.876} & 0.895 & 0.894 \\
 & Probability & \underline{0.890} & \underline{0.882} & 0.883 & \underline{0.897} & 0.875 & 0.878 & 0.883 \\
 & Predictive Entropy & 0.883 & 0.878 & 0.879 & 0.893 & 0.869 & 0.875 & 0.875 \\
 & LN Entropy & 0.884 & 0.878 & 0.879 & 0.893 & 0.873 & 0.895 & 0.881 \\
 & P(True) & 0.832 & 0.868 & \underline{0.890} & 0.886 & \underline{0.876} & 0.737 & 0.789 \\
 & Eigen Score & 0.762 & 0.742 & 0.753 & 0.767 & 0.745 & 0.802 & 0.795 \\
 & tokenSAR & 0.857 & 0.853 & 0.846 & 0.868 & 0.851 & 0.897 & 0.891 \\
 & sentenceSAR & 0.885 & 0.880 & 0.881 & 0.895 & 0.872 & 0.886 & 0.887 \\
 & SAR & 0.872 & 0.869 & 0.869 & 0.884 & 0.867 & \underline{0.914} & \underline{0.901} \\
 & Semantic Entropy & 0.768 & 0.748 & 0.757 & 0.777 & 0.765 & 0.839 & 0.832 \\
 & Ecc & 0.773 & 0.759 & 0.765 & 0.784 & 0.757 & 0.812 & 0.817 \\
 & EigV & 0.806 & 0.789 & 0.802 & 0.817 & 0.795 & 0.888 & 0.872 \\
 & Deg & 0.798 & 0.782 & 0.793 & 0.808 & 0.785 & 0.893 & 0.872 \\
 & Ours & \textbf{0.915} & \textbf{0.932} & \textbf{0.933} & \textbf{0.927} & \textbf{0.923} & \textbf{0.922} & \textbf{0.919} \\
\midrule
\multirow{14}{*}{NQ} & Perplexity & 0.748 & 0.706 & 0.728 & 0.737 & 0.729 & 0.745 & 0.749 \\
 & Probability & 0.776 & 0.747 & 0.763 & 0.774 & 0.748 & 0.729 & 0.755 \\
 & Predictive Entropy & 0.770 & 0.742 & 0.753 & 0.762 & 0.742 & 0.722 & 0.719 \\
 & LN Entropy & 0.751 & 0.716 & 0.730 & 0.744 & 0.734 & 0.720 & 0.693 \\
 & P(True) & 0.729 & 0.731 & \underline{0.785} & \underline{0.799} & \underline{0.783} & 0.609 & 0.659 \\
 & Eigen Score & 0.713 & 0.678 & 0.691 & 0.699 & 0.667 & 0.631 & 0.641 \\
 & tokenSAR & 0.732 & 0.701 & 0.714 & 0.725 & 0.714 & 0.754 & 0.750 \\
 & sentenceSAR & 0.776 & 0.747 & 0.759 & 0.768 & 0.747 & 0.732 & 0.721 \\
 & SAR & 0.776 & 0.745 & 0.755 & 0.770 & 0.754 & 0.765 & 0.729 \\
 & Semantic Entropy & 0.753 & 0.719 & 0.735 & 0.750 & 0.724 & 0.707 & 0.725 \\
 & Ecc & 0.707 & 0.697 & 0.702 & 0.714 & 0.692 & 0.665 & 0.670 \\
 & EigV & \underline{0.781} & \underline{0.753} & 0.768 & 0.780 & 0.760 & 0.759 & 0.750 \\
 & Deg & 0.758 & 0.733 & 0.748 & 0.763 & 0.737 & \underline{0.790} & \underline{0.777} \\
 & Ours & \textbf{0.839} & \textbf{0.833} & \textbf{0.840} & \textbf{0.851} & \textbf{0.836} & \textbf{0.837} & \textbf{0.823} \\
\bottomrule
\end{tabular}

}

\label{table:performance_plus_2}
\end{table*}

\begin{table*}[htbp]
\centering\caption{Experimental results on TQA and NQ with 4 additional samples. Metric: AUROC.}
\resizebox{\textwidth}{!}{
\begin{tabular}{llccccccc}
\toprule
Dataset & Method & Qwen2.5-7B & Qwen2.5-14B & Qwen2.5-32B & Qwen3-14B & Qwen3-30B-A3B & Gemma-3-12B & Gemma-3-27B \\
\midrule
\multirow{14}{*}{TQA} & Perplexity & 0.888 & 0.880 & 0.879 & 0.895 & \underline{0.876} & 0.895 & \underline{0.894} \\
 & Probability & \underline{0.890} & \underline{0.882} & 0.883 & \underline{0.897} & 0.875 & 0.878 & 0.883 \\
 & Predictive Entropy & 0.885 & 0.878 & 0.880 & 0.894 & 0.869 & 0.869 & 0.865 \\
 & LN Entropy & 0.886 & 0.879 & 0.881 & 0.895 & 0.874 & 0.885 & 0.870 \\
 & P(True) & 0.832 & 0.868 & \underline{0.890} & 0.886 & \underline{0.876} & 0.737 & 0.789 \\
 & Eigen Score & 0.816 & 0.796 & 0.810 & 0.820 & 0.798 & 0.814 & 0.805 \\
 & tokenSAR & 0.857 & 0.853 & 0.846 & 0.868 & 0.851 & 0.897 & 0.891 \\
 & sentenceSAR & 0.887 & 0.881 & 0.884 & \underline{0.897} & 0.873 & 0.879 & 0.876 \\
 & SAR & 0.880 & 0.875 & 0.877 & 0.890 & 0.874 & \underline{0.904} & 0.892 \\
 & Semantic Entropy & 0.829 & 0.807 & 0.826 & 0.839 & 0.817 & 0.859 & 0.855 \\
 & Ecc & 0.805 & 0.785 & 0.801 & 0.814 & 0.785 & 0.834 & 0.837 \\
 & EigV & 0.841 & 0.820 & 0.841 & 0.852 & 0.827 & 0.889 & 0.878 \\
 & Deg & 0.828 & 0.809 & 0.828 & 0.839 & 0.813 & 0.898 & 0.881 \\
 & Ours & \textbf{0.921} & \textbf{0.936} & \textbf{0.937} & \textbf{0.932} & \textbf{0.929} & \textbf{0.927} & \textbf{0.922} \\
\midrule
\multirow{14}{*}{NQ} & Perplexity & 0.748 & 0.706 & 0.728 & 0.737 & 0.729 & 0.745 & 0.749 \\
 & Probability & 0.776 & 0.747 & 0.763 & 0.774 & 0.748 & 0.729 & 0.755 \\
 & Predictive Entropy & 0.770 & 0.744 & 0.750 & 0.763 & 0.743 & 0.711 & 0.705 \\
 & LN Entropy & 0.753 & 0.720 & 0.732 & 0.747 & 0.736 & 0.703 & 0.673 \\
 & P(True) & 0.729 & 0.731 & 0.785 & 0.799 & 0.783 & 0.609 & 0.659 \\
 & Eigen Score & 0.737 & 0.710 & 0.722 & 0.730 & 0.702 & 0.645 & 0.645 \\
 & tokenSAR & 0.732 & 0.701 & 0.714 & 0.725 & 0.714 & 0.754 & 0.750 \\
 & sentenceSAR & 0.777 & 0.750 & 0.757 & 0.770 & 0.750 & 0.720 & 0.707 \\
 & SAR & 0.781 & 0.756 & 0.763 & 0.778 & 0.765 & 0.744 & 0.706 \\
 & Semantic Entropy & 0.790 & 0.754 & 0.779 & 0.794 & 0.766 & 0.721 & 0.715 \\
 & Ecc & 0.735 & 0.721 & 0.729 & 0.740 & 0.714 & 0.697 & 0.705 \\
 & EigV & \underline{0.808} & \underline{0.776} & \underline{0.791} & \underline{0.802} & \underline{0.786} & 0.760 & 0.745 \\
 & Deg & 0.778 & 0.753 & 0.765 & 0.777 & 0.756 & \underline{0.799} & \underline{0.783} \\
 & Ours & \textbf{0.846} & \textbf{0.836} & \textbf{0.843} & \textbf{0.853} & \textbf{0.839} & \textbf{0.842} & \textbf{0.824} \\
\bottomrule
\end{tabular}
}

\label{table:performance_plus_4}
\end{table*}

\begin{table*}[htbp]
\centering\caption{Experimental results on TQA and NQ with 8 additional samples. Metric: AUROC.}
\resizebox{\textwidth}{!}{
\begin{tabular}{llccccccc}
\toprule
Dataset & Method & Qwen2.5-7B & Qwen2.5-14B & Qwen2.5-32B & Qwen3-14B & Qwen3-30B-A3B & Gemma-3-12B & Gemma-3-27B \\
\midrule
\multirow{14}{*}{TQA} & Perplexity & 0.888 & 0.880 & 0.879 & 0.895 & 0.876 & 0.895 & \underline{0.894} \\
 & Probability & \underline{0.890} & \underline{0.882} & 0.883 & 0.897 & 0.875 & 0.878 & 0.883 \\
 & Predictive Entropy & 0.885 & 0.879 & 0.881 & 0.895 & 0.870 & 0.858 & 0.854 \\
 & LN Entropy & 0.887 & 0.880 & 0.882 & 0.896 & 0.876 & 0.870 & 0.857 \\
 & P(True) & 0.832 & 0.868 & \underline{0.890} & 0.886 & 0.876 & 0.737 & 0.789 \\
 & Eigen Score & 0.849 & 0.829 & 0.842 & 0.853 & 0.830 & 0.820 & 0.815 \\
 & tokenSAR & 0.857 & 0.853 & 0.846 & 0.868 & 0.851 & 0.897 & 0.891 \\
 & sentenceSAR & 0.888 & \underline{0.882} & 0.885 & \underline{0.898} & 0.874 & 0.868 & 0.864 \\
 & SAR & 0.884 & 0.879 & 0.882 & 0.894 & \underline{0.880} & 0.889 & 0.880 \\
 & Semantic Entropy & 0.833 & 0.817 & 0.844 & 0.844 & 0.824 & 0.851 & 0.831 \\
 & Ecc & 0.830 & 0.812 & 0.823 & 0.837 & 0.810 & 0.857 & 0.861 \\
 & EigV & 0.867 & 0.852 & 0.865 & 0.873 & 0.854 & 0.889 & 0.881 \\
 & Deg & 0.851 & 0.835 & 0.848 & 0.859 & 0.835 & \underline{0.902} & 0.888 \\
 & Ours & \textbf{0.927} & \textbf{0.936} & \textbf{0.940} & \textbf{0.935} & \textbf{0.933} & \textbf{0.930} & \textbf{0.924} \\
\midrule
\multirow{14}{*}{NQ} & Perplexity & 0.748 & 0.706 & 0.728 & 0.737 & 0.729 & 0.745 & 0.749 \\
 & Probability & 0.776 & 0.747 & 0.763 & 0.774 & 0.748 & 0.729 & 0.755 \\
 & Predictive Entropy & 0.770 & 0.745 & 0.752 & 0.763 & 0.742 & 0.697 & 0.691 \\
 & LN Entropy & 0.757 & 0.720 & 0.735 & 0.751 & 0.739 & 0.685 & 0.658 \\
 & P(True) & 0.729 & 0.731 & 0.785 & 0.799 & 0.783 & 0.609 & 0.659 \\
 & Eigen Score & 0.760 & 0.729 & 0.739 & 0.748 & 0.723 & 0.654 & 0.649 \\
 & tokenSAR & 0.732 & 0.701 & 0.714 & 0.725 & 0.714 & 0.754 & 0.750 \\
 & sentenceSAR & 0.778 & 0.751 & 0.759 & 0.770 & 0.750 & 0.705 & 0.692 \\
 & SAR & 0.789 & 0.760 & 0.769 & 0.787 & 0.773 & 0.724 & 0.688 \\
 & Semantic Entropy & 0.809 & 0.779 & 0.799 & 0.812 & 0.793 & 0.726 & 0.708 \\
 & Ecc & 0.757 & 0.732 & 0.747 & 0.755 & 0.731 & 0.741 & 0.744 \\
 & EigV & \underline{0.825} & \underline{0.793} & \underline{0.806} & \underline{0.819} & \underline{0.804} & 0.753 & 0.738 \\
 & Deg & 0.793 & 0.763 & 0.777 & 0.790 & 0.769 & \underline{0.805} & \underline{0.786} \\
 & Ours & \textbf{0.846} & \textbf{0.837} & \textbf{0.850} & \textbf{0.853} & \textbf{0.840} & \textbf{0.843} & \textbf{0.821} \\
\bottomrule
\end{tabular}
}

\label{table:performance_plus_8}
\end{table*}

\begin{table*}[htbp]
\centering\caption{Experimental results on TQA and NQ with 16 additional samples. Metric: AUROC.}
\resizebox{\textwidth}{!}{
\begin{tabular}{llccccccc}
\toprule
Dataset & Method & Qwen2.5-7B & Qwen2.5-14B & Qwen2.5-32B & Qwen3-14B & Qwen3-30B-A3B & Gemma-3-12B & Gemma-3-27B \\
\midrule
\multirow{14}{*}{TQA} & Perplexity & 0.888 & 0.880 & 0.879 & 0.895 & 0.876 & 0.895 & \underline{0.894} \\
 & Probability & \underline{0.890} & 0.882 & 0.883 & 0.897 & 0.875 & 0.878 & 0.883 \\
 & Predictive Entropy & 0.886 & 0.879 & 0.881 & 0.895 & 0.870 & 0.848 & 0.845 \\
 & LN Entropy & 0.887 & 0.880 & 0.882 & 0.898 & 0.877 & 0.856 & 0.846 \\
 & P(True) & 0.832 & 0.868 & \underline{0.890} & 0.886 & 0.876 & 0.737 & 0.789 \\
 & Eigen Score & 0.865 & 0.850 & 0.856 & 0.871 & 0.848 & 0.826 & 0.824 \\
 & tokenSAR & 0.857 & 0.853 & 0.846 & 0.868 & 0.851 & 0.897 & 0.891 \\
 & sentenceSAR & 0.889 & \underline{0.883} & 0.885 & \underline{0.899} & 0.875 & 0.856 & 0.853 \\
 & SAR & 0.886 & \underline{0.883} & 0.884 & 0.898 & \underline{0.883} & 0.874 & 0.869 \\
 & Semantic Entropy & 0.868 & 0.855 & 0.871 & 0.875 & 0.859 & 0.860 & 0.845 \\
 & Ecc & 0.841 & 0.825 & 0.835 & 0.850 & 0.824 & 0.875 & 0.871 \\
 & EigV & 0.879 & 0.867 & 0.877 & 0.887 & 0.870 & 0.884 & 0.882 \\
 & Deg & 0.862 & 0.848 & 0.859 & 0.871 & 0.849 & \underline{0.906} & 0.893 \\
 & Ours & \textbf{0.931} & \textbf{0.939} & \textbf{0.941} & \textbf{0.938} & \textbf{0.935} & \textbf{0.931} & \textbf{0.926} \\
\midrule
\multirow{14}{*}{NQ} & Perplexity & 0.748 & 0.706 & 0.728 & 0.737 & 0.729 & 0.745 & 0.749 \\
 & Probability & 0.776 & 0.747 & 0.763 & 0.774 & 0.748 & 0.729 & 0.755 \\
 & Predictive Entropy & 0.770 & 0.743 & 0.751 & 0.763 & 0.741 & 0.688 & 0.681 \\
 & LN Entropy & 0.757 & 0.720 & 0.738 & 0.752 & 0.739 & 0.673 & 0.649 \\
 & P(True) & 0.729 & 0.731 & 0.785 & 0.799 & 0.783 & 0.609 & 0.659 \\
 & Eigen Score & 0.769 & 0.735 & 0.750 & 0.756 & 0.733 & 0.661 & 0.656 \\
 & tokenSAR & 0.732 & 0.701 & 0.714 & 0.725 & 0.714 & 0.754 & 0.750 \\
 & sentenceSAR & 0.779 & 0.751 & 0.759 & 0.769 & 0.749 & 0.695 & 0.683 \\
 & SAR & 0.790 & 0.762 & 0.774 & 0.790 & 0.774 & 0.710 & 0.678 \\
 & Semantic Entropy & 0.820 & 0.796 & 0.812 & \underline{0.829} & \underline{0.810} & 0.728 & 0.710 \\
 & Ecc & 0.769 & 0.743 & 0.759 & 0.759 & 0.739 & 0.756 & 0.760 \\
 & EigV & \underline{0.834} & \underline{0.802} & \underline{0.818} & 0.827 & \underline{0.810} & 0.751 & 0.735 \\
 & Deg & 0.803 & 0.773 & 0.786 & 0.796 & 0.775 & \underline{0.808} & \underline{0.789} \\
 & Ours & \textbf{0.851} & \textbf{0.839} & \textbf{0.851} & \textbf{0.856} & \textbf{0.840} & \textbf{0.841} & \textbf{0.820} \\
\bottomrule
\end{tabular}
}

\label{table:performance_plus_16}
\end{table*}

\begin{table*}[htbp]
\centering\caption{Experimental results on XSum with 8 additional samples. Metric: AUROC.}
\resizebox{\textwidth}{!}{
\begin{tabular}{llccccccc}
\toprule
Dataset & Method & Qwen2.5-7B & Qwen2.5-14B & Qwen2.5-32B & Qwen3-14B & Qwen3-30B-A3B & Gemma-3-12B & Gemma-3-27B \\
\midrule
\multirow{14}{*}{XSum} & Perplexity & 0.681 & 0.633 & 0.670 & 0.689 & 0.596 & \underline{0.648} & \underline{0.632} \\
 & Probability & 0.553 & 0.523 & 0.493 & 0.572 & 0.521 & 0.624 & 0.616 \\
 & Predictive Entropy & 0.541 & 0.529 & 0.492 & 0.563 & 0.501 & 0.566 & 0.563 \\
 & LN Entropy & 0.659 & 0.627 & 0.638 & 0.663 & 0.558 & 0.568 & 0.570 \\
 & P(True) & 0.620 & 0.647 & \underline{0.709} & \underline{0.696} & 0.598 & 0.562 & 0.606 \\
 & Eigen Score & 0.519 & 0.431 & 0.443 & 0.623 & 0.376 & 0.524 & 0.529 \\
 & tokenSAR & 0.684 & 0.646 & 0.690 & 0.687 & 0.645 & 0.641 & 0.628 \\
 & sentenceSAR & 0.545 & 0.533 & 0.498 & 0.568 & 0.506 & 0.569 & 0.567 \\
 & SAR & \underline{0.698} & 0.676 & 0.691 & 0.692 & 0.644 & 0.596 & 0.594 \\
 & Semantic Entropy & 0.677 & 0.662 & 0.697 & 0.679 & 0.707 & 0.564 & 0.563 \\
 & Ecc & 0.624 & 0.616 & 0.615 & 0.624 & 0.632 & 0.576 & 0.579 \\
 & EigV & 0.688 & \underline{0.681} & 0.706 & 0.692 & \underline{0.709} & 0.588 & 0.585 \\
 & Deg & 0.640 & 0.637 & 0.648 & 0.647 & 0.656 & 0.591 & 0.592 \\
 & Ours & \textbf{0.751} & \textbf{0.739} & \textbf{0.797} & \textbf{0.806} & \textbf{0.781} & \textbf{0.728} & \textbf{0.713} \\
\bottomrule
\end{tabular}

}

\label{table:performance_xsum_8}
\end{table*}

\clearpage
\section{Prompts}
The generation prompts follow the design settings described in~\citet{lin2024generating} and~\citet{vashurin-etal-2025-benchmarking}. We present the generation prompts used for TQA, NQ, and XSum first, followed by the MCQ prompts employed in the self-assessment stage.

Finally, the prompt used for the LLM-as-a-Judge evaluation with GPT-4.1 is provided at the end of this section.
\begin{tcolorbox}[title=Generation Prompt for TQA, colback=white, colframe=black]
\ttfamily\small
Answer these questions:
\\
\\
Question:
\\
In Scotland a bothy/bothie is a?
\\
Answer:
\\
House
\\
\\
Question:

\{Insert Question\}

Answer:
\\
\end{tcolorbox}

\begin{tcolorbox}[title=Generation Prompt for NQ, colback=white, colframe=black]
\ttfamily\small
Answer these questions:
\\
\\
Question:
\\
who makes up the state council in russia
\\
Answer:
\\
governors and presidents
\\
\\
Question:
\\
when does real time with bill maher come back
\\
Answer:
\\
November 9, 2018
\\
\\
Question:
\\
where did the phrase american dream come from
\\
Answer:
\\
the mystique regarding frontier life
\\
\\
Question:
\\
what do you call a group of eels
\\
Answer:
\\
bed
\\
\\
Question:
\\
who wrote the score for mission impossible fallout
\\
Answer:
\\
Lorne Balfe
\\
\\
Question:
\\
\{Insert Question\}
\\
Answer:
\\
\end{tcolorbox}

\begin{tcolorbox}[title=Generation Prompt for XSum, colback=white, colframe=black]
\ttfamily\small
Here's the text and it's short one-sentence summary.
\\
\\
Text:
\\
\{Insert Text\}
\\
\\
Summary (one sentence):
\\
\\
\end{tcolorbox}

\begin{tcolorbox}[title=MCQ Prompt for TQA and NQ, colback=white, colframe=black]
\ttfamily\small
Task:
\\
Select the one correct answer to the question from the choices provided. If none of the provided choices is correct, select the final choice (\{Insert Final Label\}) None of the above.
\\
\\
Question:
\\
\{Insert Question\}
\\
\\
Choices:
\\
\{Insert Choices\}
\\
\\
Answer:
\\
The answer is (
\end{tcolorbox}

\begin{tcolorbox}[title=MCQ Prompt for XSum, colback=white, colframe=black]
\ttfamily\small
Task:
\\
Select the one correct summary for the text from the choices provided. If none of the provided choices is correct, select the final choice (\{Insert Final Label\}) None of the above.
\\
\\
Text:
\\
\{Insert Text\}
\\
\\
Choices:
\\
\{Insert Choices\}
\\
\\
Answer:
\\
The summary is (
\end{tcolorbox}

\begin{tcolorbox}[title=LLM-as-a-Judge Prompt for TQA and NQ, colback=white, colframe=black]
\ttfamily\small
\textbf{System Message}:
\\
\\
\# Task
\\
\\
Evaluate whether the proposed answer 
to the question is correct based
on real-world factual knowledge. 
Reference answers are provided to
assist in your evaluation.
\\
\\
\# Output
\\
\\
Respond strictly with a single token:

- `True' if the proposed answer is 
correct.

- `False' if the proposed answer is 
incorrect or only partially correct.
\\
\\
\textbf{User Message}:
\\
\\
Question:

\{Insert Question\}
\\
\\
Reference Answer(s):

\{Insert Reference Answers\}
\\
\\
Proposed Answer:

\{Insert Proposed Answer\}
\\
\\
True/False:
\end{tcolorbox}
\newpage
\begin{tcolorbox}[title=LLM-as-a-Judge Prompt for XSum, colback=white, colframe=black]
\ttfamily\small
\textbf{System Message}:
\\
\\
\# Task
\\
\\
Evaluate whether the proposed summary 
is correct based on the original text. 
A reference summary is provided to assist 
in your evaluation.
\\
\\
\# Output
\\
\\
Respond strictly with a single token:

- `True' if the proposed summary is 
accurate and faithful to the original text.

- `False' if the proposed summary is 
inaccurate or misleading.
\\
\\
\textbf{User Message}:
\\
\\
Original Text:

\{Insert Text\}
\\
\\
Reference Summary:

\{Insert Reference Summary\}
\\
\\
Reference Summary:

\{Insert Proposed Summary\}
\\
\\
True/False:
\end{tcolorbox}

\end{document}